\documentclass[runningheads]{llncs}
\usepackage[T1]{fontenc}
\usepackage{graphicx}
\usepackage{booktabs}
\usepackage[misc]{ifsym}

\usepackage{mwe}

\usepackage{amssymb,amsmath,array}
\usepackage{tikz}
\usetikzlibrary{arrows,shapes}
\usetikzlibrary{patterns}
\usetikzlibrary{decorations.pathreplacing, patterns}
\usepackage{tabularx}
\usepackage{multicol}
\usepackage{multirow}
\usepackage{algpseudocode}
\usepackage{hyperref}
\usepackage{listings}
\usepackage[small]{caption}
\usepackage{algorithm}
\usepackage[switch]{lineno}
\usepackage{bbm}

\usepackage{multirow}%
\usepackage{xcolor}
\usepackage{multicol}

\newfloat{algorithm}{t}{lop}
\algnewcommand\algorithmichyperparams{\textbf{Hyperparameters:}}
\algnewcommand\AHyperparams{\item[\algorithmichyperparams]}

\algnewcommand\algorithmicparams{\textbf{Parameters:}}
\algnewcommand\AParams{\item[\algorithmicparams]}

\algnewcommand\algorithmicinput{\textbf{Input:}}
\algnewcommand\algorithmicoutput{\textbf{Output:}}
\algnewcommand\Input{\item[\algorithmicinput]}
\algnewcommand\Output{\item[\algorithmicoutput]}
\algnewcommand\algorithmicforeach{\textbf{for each}}
\algdef{S}[FOR]{ForEach}[1]{\algorithmicforeach\ #1\ \algorithmicdo}

\begin{document}

\title{Open World Autoencoding Drift Detection with Novel Class Recognition in Tabular Non-stationary Data Streams}
\titlerunning{Open World Autoencoding Drift Detection}

\author{Joanna Komorniczak}

\authorrunning{J. Komorniczak}

\institute{\textit{Department of Systems and Computer Networks},\\ Wrocław University of Science and Technology}

\maketitle              

\begin{abstract}
Data stream processing has become a landmark in modern machine learning applications, with \textit{concept drifts} and \textit{novel class appearances} posing the primary challenges faced by sophisticated recognition methods. This work proposes an unsupervised concept drift detection method that identifies shifts in known class distributions based on the reconstruction errors of an autoencoder, while also enabling the recognition of novel class samples through density estimation of a \textit{proxy} representation of samples. Using \textit{mirrored} autoencoders allows for independent incremental adaptation to changing problem distributions for the two considered tasks, resulting in continuous adjustment to evolving concepts and reliable recognition of unknown samples. Conducted experiments used a diverse set of synthetic tabular data streams, where both concept drifts and the emergence of novelties were observed. The results show that the proposed approach is competitive with current state-of-the-art unsupervised drift detectors and novelty classifiers.

\keywords{data stream  \and concept drift \and novelty detection \and concept drift detection}
\end{abstract}

\section{Introduction}

\begin{figure}[!htb]
    \centering
    \includegraphics[width=0.8\linewidth, trim={2cm 0 3cm 0}, clip]{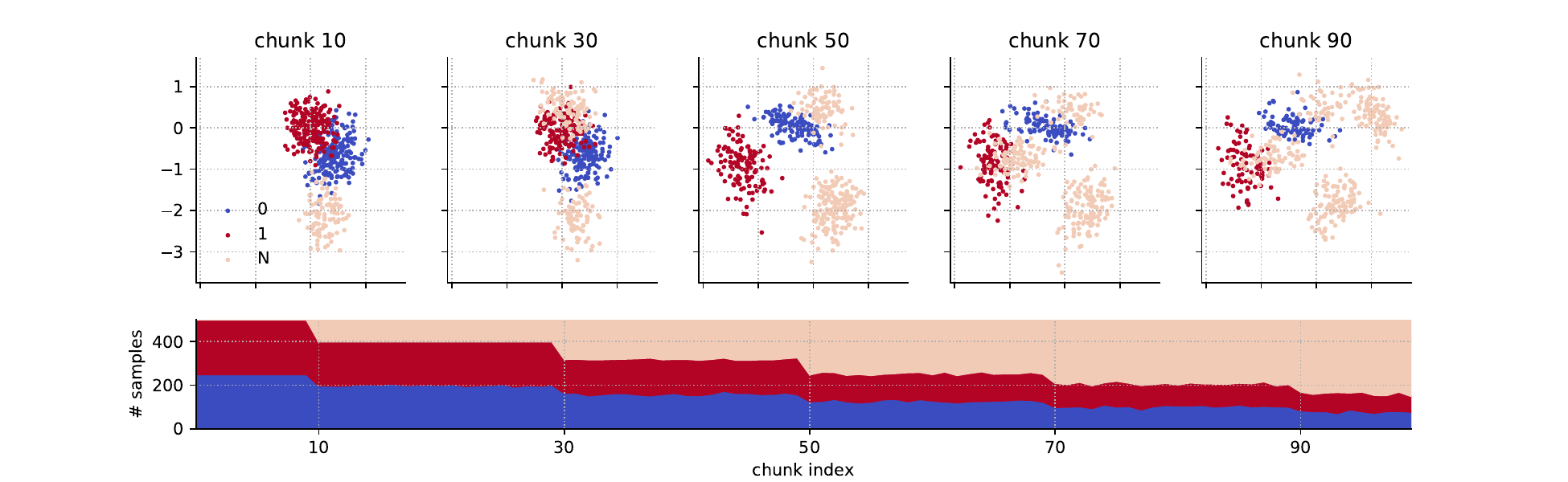}
    \caption{An example of a non-stationary data stream, where over 100 chunks, five novel classes appeared (beige markers). The known class distribution (blue and red markers) shifted due to a concept drift in the 50$^{th}$ data chunk. The bottom plot shows the proportions of classes in a given timestamp.}
    \label{fig:vis_stream}
\end{figure}

Fueled by the growing popularity of digital media, today's world generates, processes, and stores immense amounts of data that align well with the paradigm of \textit{continuous} and \textit{dynamic} data stream processing \cite{de2016iot}. The modern machine learning systems employed for this purpose face various challenges, most often related to the non-stationarity of data, which may manifest as concept drift \cite{agrahari2022concept}. Since many systems process samples in real-time, the availability of labels for model training and adaptation may be limited \cite{gomes2022survey}. Moreover, some instances may represent previously unseen object categories -- a result of the emergence of novel, unknown observations \cite{gaudreault2024systematic}. In an \textit{open world} \cite{scheirer2012toward}, where the training set represents only part of reality, methods should be ready to signal shifts in sample distribution but also to mark instances that may represent previously unseen categories of objects. Figure \ref{fig:vis_stream} presents a data stream with the appearance of novel categories and a concept drift that affects the distribution of the known classes.

\paragraph{Concept drift detection}

Concept drift detection is an essential task in many systems operating in a data stream setting, as recognition methods may degrade due to a shift in data distribution \cite{page1954continuous}. The first generation of concept drift detectors directly employed the fact that the recognition quality of a classifier lowers as a result of a \textit{real} change in posterior probability $P(Y|X)$ distribution \cite{gama2004learning,bifet2007learning}. More recent methods focus on analyzing the distribution of data -- allowing for implicit detection \cite{gozuaccik2021concept} -- and therefore for the recognition of \textit{virtual} drifts in covariate distribution $P(X)$ that do not yet lead to a loss of classifier performance. Among the implicit methods, unsupervised methods represent an important category of drift detection approaches \cite{lukats2024}. These methods are particularly valuable in applications where the arrival of labels may be delayed, or their acquisition may be costly \cite{gomes2022survey}.

Implicit drift detectors can still utilize the classifier, which is done by the Margin Density Drift Detector (\textsc{md3}) \cite{sethi2015don}, but only to study the density of samples near the decision boundary. The Discriminative Drift Detector (\textsc{d3}) \cite{gozuaccik2019unsupervised} uses a linear regression model to check if the distinction between the current and historical distributions represents an easy task. Other approaches, such as \textsc{ocdd} \cite{gozuaccik2021concept}, employ one-class classifiers to detect changes based on the number of reported outliers. Among the distribution-based approaches, there are also drift detectors that use statistical tests to compare two windows of observations. The Kolmogorov-Smirnov test has been used for that purpose \cite{porwik2022detection}. Some methods, like the \textsc{padd} \cite{komorniczak2024unsupervised}, employ repeated T-Tests with randomly sampled instances. In this approach, high-dimensional samples are transformed into a lower-dimensional space on the outputs of a neural network. 
Reduction of data dimensionality can also be performed using autoencoders. One method that employs this idea is the Autoencoder Drift Detector (\textsc{add}) \cite{menon2020concept}, which was used to detect phishing attacks. Another case study used autoencoders for \textsc{eeg} analysis \cite{khadimallah2024autoencoder}, employing a separate model for each class in the data stream. While most methods measure the reconstruction error cost function, some also rely on the cross-entropy \cite{jaworski2020concept}. The comparison of recent and historical data in many methods also employs statistical hypothesis testing, e.g., the Page-Hinckley test \cite{zhan2023unsupervised}, or the Mann-Whitney U Test \cite{li2023autoencoder}. 

In one interesting work where both concept drifts and novelties were detected \cite{shang2025novelty}, the authors used the \textit{radial base distance} to detect significant distribution changes. While the work considers the benefits of dimensionality reduction and mentions PCA as a tool for that purpose, the method itself analyzes the original high-dimensional samples.

\paragraph{Novelty detection}

The emergence of a novel class -- even in an unsupervised setting, where the labels of the novel class have not yet arrived -- affects the data covariate probability $P(X)$. It can therefore be interpreted as a specific case of concept drift. In contrast to canonical concept drift, which affects only known classes, this type of change will not be recognized by methods explicitly monitoring the errors of classifiers \cite{shang2025novelty} until the classifier is adapted to the evolving recognition problem.

While autoencoders find applications in drift and outlier detection, most existing novelty detection methods directly analyze the neighborhood of original samples or the distances between clusters \cite{faria2016novelty}. One of the first methods for novelty detection, \textsc{olindda} \cite{spinosa2009novelty}, relies on clusters formed using the k-means algorithm. ECSMiner \cite{masud2010classification} employs an ensemble of classifiers to recognize among known classes. To detect novelty, it clusters the samples stored in short-term memory and analyzes the distances between known and unknown examples using a measure named \emph{q}-\textsc{nsc}. Another popular method -- \textsc{minas} \cite{de2016minas}, similarly to \textsc{olindda}, relies on clustering. To detect novelty, it thresholds the distance among centroids of clusters that represent unknown and known samples. We note that the purpose of these sophisticated novelty detectors is not only to mark the unknowns but also to recognize \textit{novelty patterns} and even distinguish between more than one novelty in a given data chunk \cite{faria2016novelty}. These opportunities come with a cost in terms of the method's memory complexity, as the original feature vector needs to be stored in a buffer.

When novelty detection shifts from tabular to unstructured data, many concepts overlap with open-set recognition \cite{gaudreault2024systematic}. In that case, the existing approaches cannot effectively store the original features of unknowns, therefore, methods often rely on object embeddings \cite{zhou2021learning}. One of the methods for \textit{open set recognition} that relies on autoencoding is \textsc{c2ae} \cite{oza2019c2ae}, which trains the encoder and decoder independently to allow for classification among known classes.

\paragraph{Contribution}

This article focuses on the complex task of processing non-stationary data streams, where both concept drift and unknown samples are present. The main contributions of the presented work are as follows:
\begin{itemize}
    \item The proposition of the \textsc{owadd} method -- an unsupervised drift detector analyzing the reconstruction errors of an autoencoder and the replicated T-Test to recognize significant changes in the data distribution.
    \item The use of reconstruction errors to recognize novel class instances based on density estimation with \textsc{kde} integrated with the proposed detection approach.
    \item A shift from the analysis of the entire set of features to just a one-dimensional \textit{proxy} vector of reconstruction errors, reducing the memory requirements of the method.
    \item The evaluation of the proposed methodology on a diverse set of nearly 2,000 data streams, while also considering the imbalanced nature of the novelty recognition task, where the proportion of known and unknown class samples changes over processing time.
\end{itemize}

\section{Method}

This article presents an \textit{Open World Autoencoding Drift Detector} (\textsc{owadd}). The method employs two autoencoding models with fully connected layers, studying the reconstruction errors of the samples to detect concept drift and identify unknown samples. The method is fully unsupervised, hence it does not depend on labels, even in the initial \textit{offline} phase, making it suitable for high-velocity data streams where data annotation may be costly or burdened with a delay. The method's ability to recognize unknown samples makes it valuable in an open-world setting \cite{scheirer2012toward}, where modern machine learning methods must face unexpected events provided to recognition systems.

\begin{algorithm}[h]
\scriptsize
\caption{\textit{Open World Autoencoding Drift Detector}}
\label{alg:alg}
\begin{algorithmic}[1]
\Input
\Statex $\mathcal{DS} = [\mathcal{X}_1, \mathcal{X}_2, \ldots ]$ -- unlabeled data stream

\AHyperparams
\Statex $\mathcal{A}$ -- fully connected autoencoder,
$e$ -- number of training epochs,
$r$ -- number of statistical test replications,
$N$ -- sample size,
$\theta$ -- drift detection threshold,
$\delta$ -- novelty recognition threshold
\AParams
\Statex $d$ -- drift indicator,
$\mathcal{B}$ -- reference buffer of reconstruction errors, 
$\mathcal{A}_{KC}$ -- autoencoder for known-class recognition, 
$\mathcal{K}$ -- kernel density estimation model

\Output
\Statex $[(d_1, \mathcal{Y}_1), (d_2, \mathcal{Y}_2), \ldots ]$ -- drift indicators $d_n$ and labels of known and unknown samples $\mathcal{Y}_n$

\vspace{1em}

\State $\mathcal{B} \gets \emptyset$
\ForAll{$\mathcal{X}_n \in \mathcal{DS}$}

    \If{$n$ = 0}
        \State train $\mathcal{A}$ on $\mathcal{X}_n$ for $e$ epochs
        {\color{gray}\Comment{training with first data batch}}
        \State $\mathcal{A}_{KC} \gets$ copy of $\mathcal{A}$
    \EndIf

    \State $L_n \gets [ \sum|x - \mathcal{A}(x)|$ for $ x \in \mathcal{X}_n ]$
    {\color{gray}\Comment{collect reconstruction errors}}

    \If{$\mathcal{B}$ is not full}
        \State $\mathcal{B} \gets \mathcal{B} \cup L_n$
        {\color{gray}\Comment{extend buffer}}

        \State $d \gets$ \textit{False}
        \State $\mathcal{Y}_n \gets$ \Call{RecognizeUnknown}{$\mathcal{X}_n$}
        \State \Return $(d, \mathcal{Y}_n)$
    \EndIf

    \State $p \gets 0$
    {\color{gray}\Comment{initialize counter}}

    \For{$i = 1$ to $r$}
    \State $R_i \gets$ random $N$ errors from $\mathcal{B}$
    \State $C_i \gets$ random $N$ errors from $L_n$
        \State perform one-sided \textit{T-test}: $H_0: R_i\geq C_i$
        \If{$p\text{-value} < 0.05$}
            \State $p \gets p + 1$
            {\color{gray}\Comment{increment counter when difference is significant}}

        \EndIf
    \EndFor

    \If{$p / r > \theta$}
    {\color{gray}\Comment{evaluate ratio of positive outcomes}}
        \State $d \gets$ \textit{True}
        \State train $\mathcal{A}$ on $\mathcal{X}_n$ for $e$ epochs
        \State $\mathcal{B} \gets \emptyset$
        {\color{gray}\Comment{signal drift}}
    \Else
        \State $d \gets$ \textit{False}
    \EndIf

    \State $\mathcal{Y}_n \gets$ \Call{RecognizeUnknown}{$\mathcal{X}_n$}
    {\color{gray}\Comment{mark unknown samples}}

    \State \Return $(d, \mathcal{Y}_n)$

\EndFor
\end{algorithmic}

\end{algorithm}

Algorithm \ref{alg:alg} presents the operation of the proposed approach. The method processes an unlabeled data stream $\mathcal{DS} = [\mathcal{X}_1, \mathcal{X}_2, \ldots ]$ in batches $\mathcal{X}_n$. The result of a procedure for each data batch are (a) the boolean drift indicator $d_n$ and (b) the labels describing known and unknown samples $\mathcal{Y}_n$. The method employs a fully connected autoencoder $\mathcal{A}$, which is trained on the first data batch over $e$ epochs. This first batch serves as an \emph{offline} phase. Other hyperparameters of the proposed approach are the number of statistical test replications $r$, the sample size $N$, and thresholds for drift detection $\theta$ and novelty recognition $\delta$. The number of test replications and the sample size were set to $15$ and $30$, respectively, and remained invariant. The method stores the reference reconstruction errors in a buffer of size $1000$. It is important to note that the buffer is not intended for samples of potentially high dimensionality, but only for one-dimensional reconstruction errors, which significantly limits the memory complexity of the method.

At the beginning of a procedure, the empty buffer $\mathcal{B}$ is initialized. The first data stream chunk is used to train the autoencoder $\mathcal{A}$, responsible for drift detection. A \emph{mirror} of a trained autoencoder $\mathcal{A}_{KC}$ is created, which will be responsible for the recognition of novel class samples. The weights of this additional model are updated only when explicitly requested, i.e., when the appearance of novelty is confirmed, and the previously unknown classes should no longer be viewed as novelties. In contrast, the weights of $\mathcal{A}$ will be incrementally updated with each drift detection to allow adjustment to the reference distribution for continuous future change recognition (lines 2:6 of the pseudocode). 
After initial autoencoder training, the method computes the reconstruction errors $L_n$ for each sample in a data chunk (line 7). If the reference buffer is not full, errors are added to the buffer $\mathcal{B}$, and the drift detection procedure is not yet performed.
Once the reference buffer is full, $r$ replications of a one-sided T-test are performed, each time comparing the equal-sized samples of size $N$ randomly drawn from the reference buffer $\mathcal{B}$ and the errors of the current chunk $L_n$. According to the number of tests that have shown a significant difference under $\alpha=0.05$ and the drift detection threshold $\theta$, concept drift is detected. This procedure is described in lines 14:24. In the case of drift detection, the weights of $\mathcal{A}$ are updated, and the buffer is reset (lines 25:26). For each data chunk, the additional procedure that recognizes novel class samples is called, resulting in a vector $\mathcal{Y}_n$ describing known (positive) and unknown (negative) class predictions. This function is described in Algorithm \ref{alg:rec}.

\begin{algorithm}[h]
\scriptsize
\caption{Recognition of novel class samples}
\label{alg:rec}
\begin{algorithmic}[1]

\Function{RecognizeUnknown}{$\mathcal{X}$}
    \State $E \gets [ \sum|x - \mathcal{A}_{KC}(x)|$ for $x \in \mathcal{X}]$
    {\color{gray}\Comment{collect reconstruction errors}}
    \If{$\mathcal{K}$ is undefined}
        \State $\mathcal{K}\gets$ fit $\mathcal{K}$ using $E$
        {\color{gray}\Comment{fit estimator}}
    \EndIf
    \State  $S \gets$ density scores of $\mathcal{K}(E)$
    \State $\mathcal{Y} \gets \mathbbm{1} [S \geq \delta]$
    {\color{gray}\Comment{recognize unknown instances}}
    \State \Return $\mathcal{Y}$
\EndFunction
\end{algorithmic}

\end{algorithm}

The function uses $\mathcal{A}_{KC}$ to calculate the reconstruction errors $E$ (line 2). This additional model is needed to immunize $\mathcal{A}$, incrementally trained after each drift detection, since not all concept drifts are associated with novel class appearances. \textsc{Owadd} offers the possibility to additionally fit the $\mathcal{A}_{KC}$ model once the novelty is confirmed and the reference distribution of known classes changes. The novelty recognition procedure employs a \textsc{kde} density estimation model $\mathcal{K}$, fitted to the initial distribution of known-class errors (lines 3:5). The use of a one-dimensional \textit{proxy} makes \textsc{kde} resistant to the curse of dimensionality \cite{verleysen2005curse}, which heavily impacts the density estimation task \cite{diakonikolas2019robust}. This is yet another strength of the \textsc{owadd} method that comes from shifting the focus from the analysis of the entire feature vector of each sample $x_i \in \mathcal{X}_n$ to its one-dimensional reconstruction error. To recognize novel samples, the function employs the novelty threshold $\delta$ on the density scores obtained for current errors $E$ using $\mathcal{K}$. The samples whose scores lie above the threshold are labeled as known (1), while the remaining ones are labeled as unknown (0), as described in lines 6:7. The complete procedure results in the drift indicator $d_n$ and the labels $\mathcal{Y}_n$ returned for each data chunk $n$.

\begin{figure}[!b]
    \centering
    \includegraphics[width=0.9\linewidth]{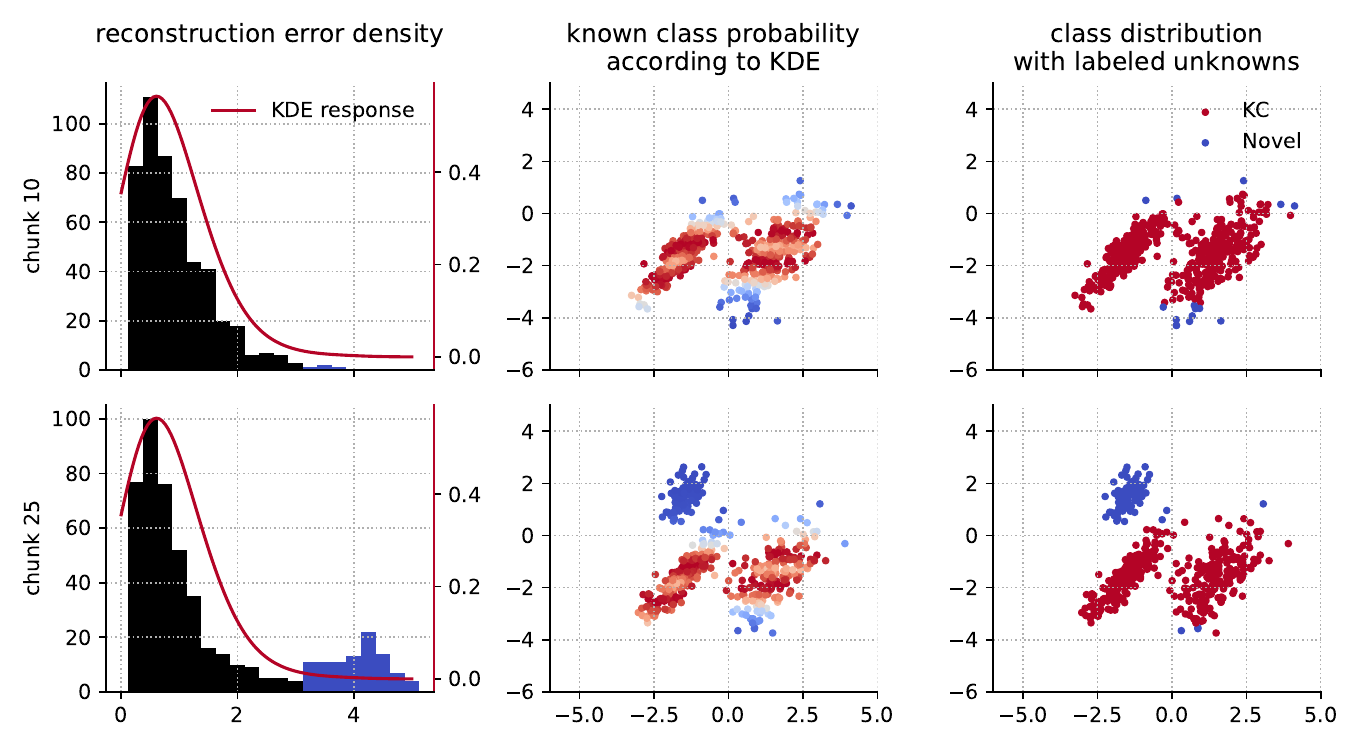}
    \caption{The initial class distribution in 10$^{th}$ data chunk (first row) and distribution after concept drift with novel class appearance in 25$^{th}$ chunk (second row). The recognition of unknown samples (last column) is based on the estimated density of reconstruction errors.}
    \label{fig:example-owadd}
\end{figure}

Figure \ref{fig:example-owadd} presents the internal components of the method using an exemplary planar data stream, where the concept drift is associated with the emergence of a new class in the 25$^{th}$ data chunk. The first row shows the distribution of the problem in the 10$^{th}$ batch, which is not yet impacted by concept drift. The first column presents the distribution of reconstruction errors for those two time instants with \textsc{kde} fitted to the distribution. One can notice an increase in the number of significant reconstruction errors (marked in blue), which is a sign of concept drift. The second column presents the original distribution of data, coloring the samples according to their scores from the \textsc{kde} model. Low density is associated with the blue color. The last column indicates whether the samples are labeled as known (red) or novel (blue) according to the novelty threshold $\delta$. We note that this simple two-dimensional example serves as a visualization, while the method is intended for data of higher dimensionality.

The method benefits from the use of reconstruction errors that limit the dimensionality from the initial number of features to just one value per sample. The reconstruction errors allow for both detecting concept drifts and recognizing novel classes. The incremental training of the autoencoder $\mathcal{A}$ after the drift allows for automatic adjustment of the reference distribution, resulting in effective and fully unsupervised drift detection.

\section{Experiment design}

A series of experiments was conducted to study the behavior of the proposed approach and to compare it with state-of-the-art methods in the tasks of (a) unsupervised concept drift detection and (b) recognition of novel instances.

\subsubsection{Data streams with concept drift and class novelties}

The method was designed for data streams that exhibit both concept drift and class novelty. Similar to existing research on these two phenomena \cite{cejnek2018concept,masud2010classification}, we focused on evaluating synthetic data, where the number of events (concept drifts and novelties) and other data stream parameters could be configured, including the proportion of novel class instances and class separation. The research utilized 195 diverse data streams, with ten replications of specific configurations, thereby enabling a thorough statistical analysis of the obtained results. We also note that an unambiguous ground-truth of concept change occurrence is necessary to reliably evaluate the concept drift detection task \cite{bifet2017classifier}, making the use of synthetic data highly favorable.

The distribution of known classes shifted due to concept drift, and previously unknown clusters of observations appeared as a result of novelty \cite{komorniczak2025synthetic}. Each of the generated data streams consisted of 40,000 instances, divided into chunks of size 200. The problem was described by 50 numeric features. The initial number of known classes was set to two and grew throughout the processing to up to 11 classes, with various proportions of novel instances in each data chunk. Table~\ref{tab:generator} shows the complete configuration of the data stream generation procedure. 

\begin{table}[!htb]
    \caption{Hyperparameters used in the data streams generation procedure}
    \centering
    \setlength{\tabcolsep}{13pt}
    \renewcommand{\arraystretch}{1.2}
    \tiny
    \begin{tabular}{l|l|l}
    \toprule
         \textsc{hyperparameter} &  \textsc{drift detection task} & \textsc{novelty recognition task} \\
         \midrule
         number of chunks & 200 & 200 \\
         chunk size & 200 & 200 \\
         number of known classes & 2 & 2 \\
         problem dimensionality & 50 features & 50 features \\ \midrule
         number of concept drifts & 0, 2, 4 & 0 \\
         number of novel classes & 3, 5, 7 & 5, 7, 9 \\
         proportion of novel class & 0.05, 0.1, 0.2, 0.3 & 0.2 \\
         class separation & 1.0, 1.5, 2.0, 2.5, 3.0 & 1.0, 1.5, 2.0, 2.5, 3.0 \\ \midrule
        replications & 10 & 10 \\
    \bottomrule
    \end{tabular}
    \label{tab:generator}
\end{table}

The \textit{class separation} hyperparameter describes the degree of separation between clusters of specific classes. Both tasks of concept drift detection and novelty recognition should be easier when the class separation and the proportion of novel class instances are significant. Under each configuration, the generation procedure was repeated 10 times with various random states, resulting in the evaluation of 1950 data streams and allowing for the statistical analysis of the obtained results.

\subsubsection{Evaluation metrics}

The experiments focused on the two main challenges related to non-stationary data stream processing, each evaluated with specific metrics and compared with specific state-of-the-art reference methods. In the task of drift detection, we used three \textit{drift detection error measures} \cite{komorniczak2022statistical}: \textit{D1} -- the distances between the detections signaled by the method and the nearest concept drift; \textit{D2} -- the distances between the actual drift occurrence and the nearest detection of the method; \textit{R} -- the adjusted proportion of drifts to the number of concept detections.
These measures analyze the timestamps of actual events occurring in the stream, treating the appearance of novelty as a specific case of a concept drift and resulting in a thorough evaluation under three equally important and complementary criteria.

In the novelty recognition experiment, we evaluated the ability of methods to distinguish between known and unknown (novel) instances. Hence, this experiment considered binary classification metrics, where known instances were labeled as positive and unknown instances as negative. Since the proportion of known and unknown instances varied across the data stream, the evaluation considered the class imbalance, resulting in the assessment of three metrics: \textit{balanced accuracy} -- a symmetric measure \cite{brzezinski2019dynamics}, calculated as an average of recall and specificity in a binary classification case; \textit{recall} -- describes what share of positive (known) objects were recognized correctly; \textit{specificity} -- describes what share of negative (unknown) objects were recognized correctly. Such a three-criterion evaluation enables a thorough assessment of the method's performance in an imbalanced setting, where the proportions between the knowns and unknowns vary over time.

\subsubsection{Goals of experiments}

The experiments compared \textsc{owadd} with reference drift detection and novelty recognition approaches. The hyperparameters of all methods were optimized for the considered data streams prior to performing the research. As a result, the \textsc{owadd} method used the autoencoder with three layers of 10 neurons each, whose weights were optimized over 400 epochs. The detection threshold was set to 0.3, and the novelty threshold to 0.02.

\paragraph{Unsupervised drift detection}
 The first experiment compared the proposed approach with six unsupervised drift detectors \cite{lukats2024}, designed for tabular and structured data. The reference methods, and their hyperparameter configurations were as follows:
 \begin{itemize}
     \item \textit{Discriminative Drift Detector} (\textsc{d3}) \cite{gozuaccik2019unsupervised} with a threshold of 0.8.
     \item \textit{Kolmogorov Smirnov Drift Detector} (\textsc{ksdd}) with a window size of 200 instances and a threshold of 0.005. This method analyzed the mean across the feature values.
     \item \textit{One Class Drift Detector} (\textsc{ocdd}) \cite{gozuaccik2021concept} with a threshold of 0.75.
     \item \textit{Parallel Activations Drift Detector} (\textsc{padd}) \cite{komorniczak2024unsupervised} with a threshold of 0.1.
     \item \textit{Centroid Distance Drift Detector} (\textsc{cddd}) \cite{klikowski2022concept} with a filter size of 3.
     \item \textit{Margin Density Drift Detector} (\textsc{md3}) \cite{sethi2015don} with a sigma of 0.1. Since the initial implementation of this method requires the true class labels at algorithm initialization and after the drift occurrence, these were estimated using the \textit{k-means} clustering algorithm to allow for a comparison in a fully unsupervised setting.
 \end{itemize}

 The novelties that appeared in the data streams were also considered a concept drift. Since the distribution shift in the case of novel class appearance is more subtle, we expect those events to pose a greater challenge to the methods, especially when the proportion of novel class instances is small (in extreme cases, as low as 5\%).

\paragraph{Novelty recognition}

The second experiment focused on the ability of methods to recognize novel samples. In this task, the number of considered classes increased throughout the data stream processing period. We evaluated the proposed \textsc{owadd} with four reference methods. The reference methods and their configuration were as follows:
\begin{itemize}
    \item ECSMiner \cite{masud2010classification} with K=10 and an ensemble size of 6.
    \item \textsc{minas} \cite{de2016minas} with $k=3$, a threshold factor of 1.1, and a window size of 1000.
    \item One-class Novelty Detector (\textsc{ocnd}) -- a method that uses a one-class \textsc{svm} classifier to recognize known class samples. A similar model is employed in the \textsc{ocdd}, which signals drift based on the number of outliers identified by the classifier. The method uses the \textsc{svm} with an \textsc{rbf} kernel and $nu=0.5$.
    \item Confidence Novelty Detector -- a baseline employing the confidence of the \textsc{mlp} Classifier. The samples will be labeled unknown if the classification confidence (maximum class probability) lies below the threshold of 0.9.
\end{itemize}

All reference classifiers have access to the true class labels in the first data stream chunk. However, since the main considered scenario was unsupervised data stream processing, the models were not further updated during stream processing and relied solely on the initial training. The proposed \textsc{owadd} also recognized unknowns according to the distribution in the \textit{offline} phase, but discarded the true class labels.

\section{Experiment results}


\subsubsection{Unsupervised drift detection}

\begin{figure}
    \centering
    \includegraphics[width=\linewidth]{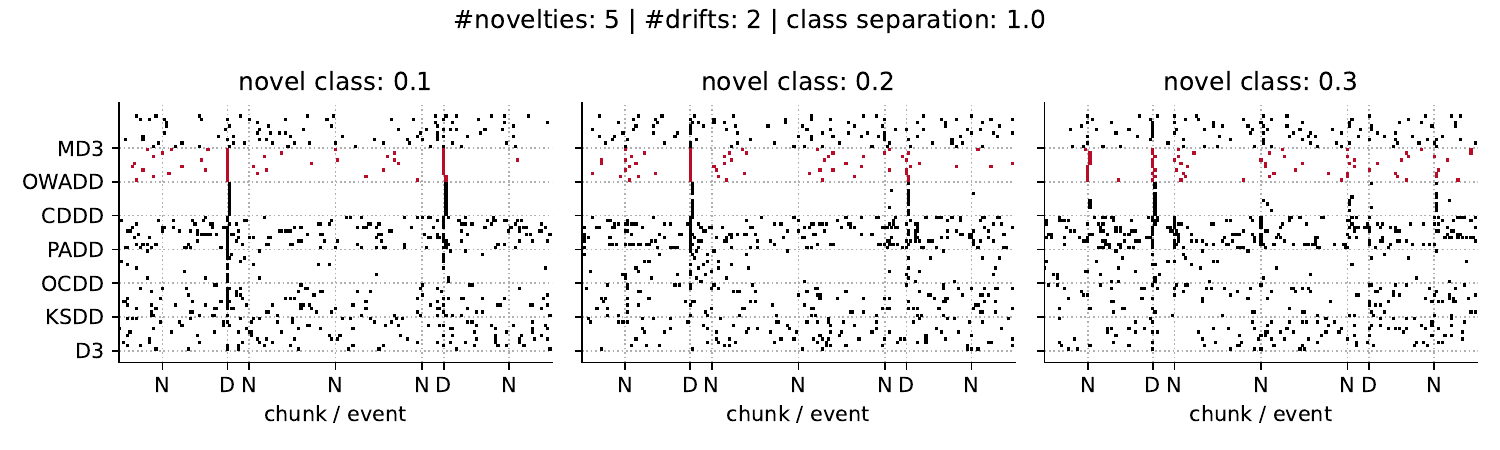}
    \caption{Detection moments (points) over 10 stream generation replications with studied methods for a data stream with five novel classes and two concept drifts (marked on the x-axis). The proposed approach is shown in red.}
    \label{fig:e3_example}
\end{figure}

The first experiment evaluated the task of concept drift detection and compared the proposed approach to reference unsupervised drift detectors under various streaming scenarios. Figure \ref{fig:e3_example} shows the results across 10 experiment replications for a data stream with five novel classes and two concept drifts. The three subfigures illustrate the detection moments for various proportions of novel classes within a given chunk. In the results presented on the left, 10\% of samples in a given chunk described a new class after novelty emergence, 20\% in the streams presented in the center, and 30\% on the right. As expected, the novelty is easier to spot in the case of a higher proportion of novel classes. Concept drifts, indicated with $D$ on the horizontal axes, are also easier to detect compared to novelties (indicated with $N$).
 
\begin{figure}
    \centering
    \includegraphics[width=\linewidth]{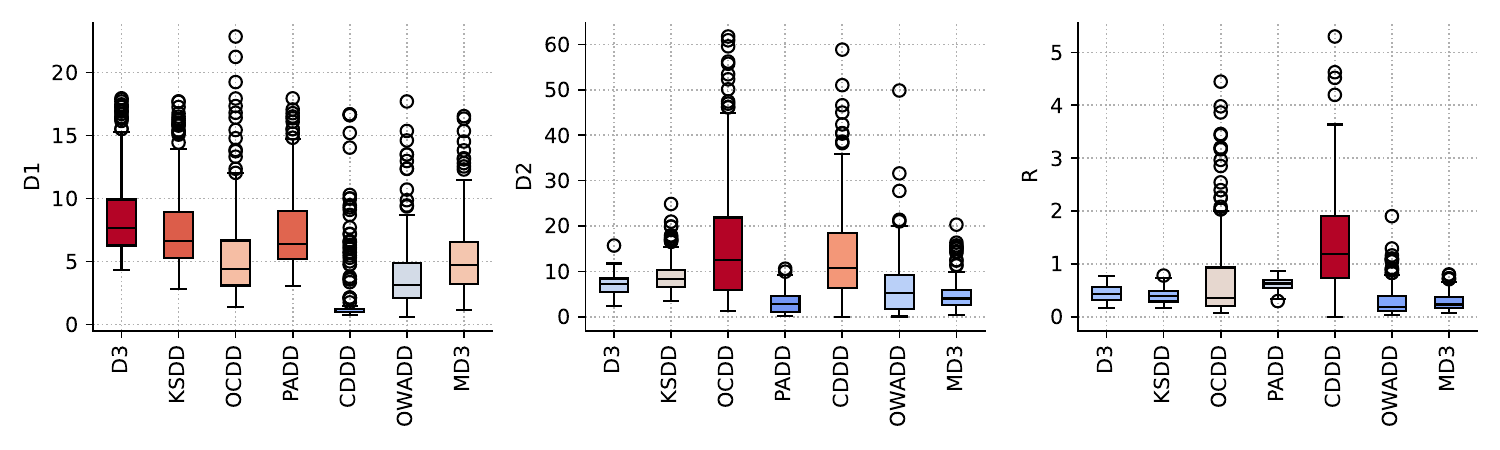}
    \caption{The distribution of results for evaluated drift detectors across three drift detection error measures: D1 (left), D2 (center), and R (right).}
    \label{fig:e3_dde_box}
\end{figure}

Figure \ref{fig:e3_dde_box} presents the distribution of the complete set of results in three \textit{drift detection error measures}. The lower error indicates better performance of the method. Each box in the figure shows the distribution of errors obtained by a specific method, with the color of the box representing the average value of the evaluation measure. Low errors are associated with the blue color, while high errors with red. The results highlight that these three criteria should be treated as complementary, since often a good result in one measure is not reflected in others --  e.g., \textsc{cddd}, which has the lowest errors in D1, suffers from high errors in D2 and R, resulting from its inability to detect novelties, as also shown in Figure \ref{fig:e3_example}.

\begin{figure}[!htb]
    \centering
    \includegraphics[width=0.32\linewidth, trim={1cm 0.2cm 1cm 0.2cm}, clip]{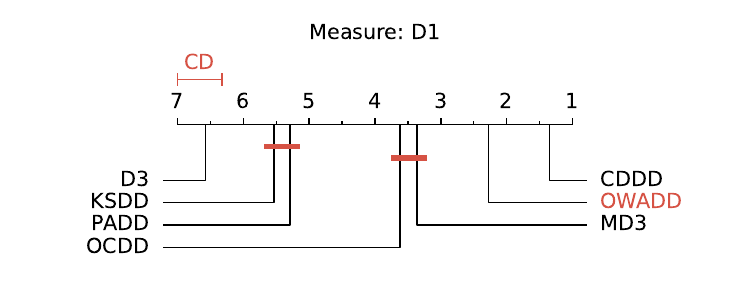}
    \includegraphics[width=0.32\linewidth, trim={1cm 0.2cm 1cm 0.2cm}, clip]{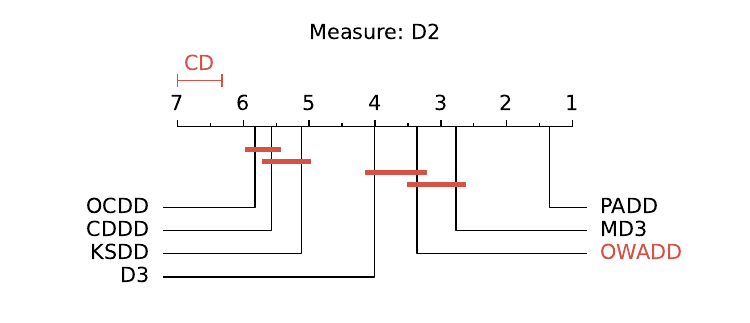}
    \includegraphics[width=0.32\linewidth, trim={1cm 0.2cm 1cm 0.2cm}, clip]{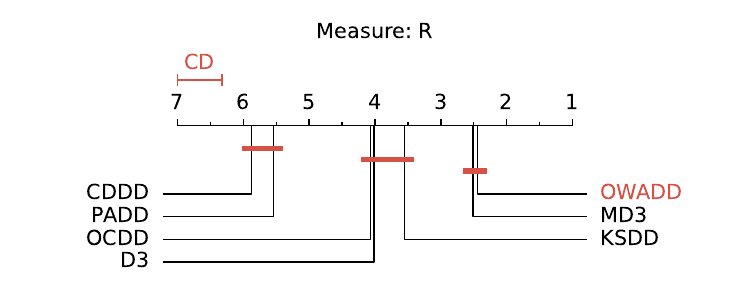}
    \caption{Critical difference diagrams for results of drift detection experiment. The methods on the right end of the axes achieve the best results. Results that are statistically related (their difference is less than CD) are connected with the red line.}
    \label{fig:e3_cd}
\end{figure}

The detailed results for the selected stream types are also presented in Table \ref{tab:e3_all}, with an average rank across all considered data stream types. The column of data stream type describes the number of novelties (N) and the separation factor of class clusters (S). If the results of the method were statistically significant, they have been emphasized in bold. The statistical analysis was conducted using the non-parametric Friedman test with $\alpha = 0.05$. To evaluate the relationship between the ranks, we used the post-hoc Nemenyi test \cite{demvsar2006statistical}. The results for all data streams are also graphically presented in Figure \ref{fig:e3_cd}.

\begin{table}[!b]
\caption{Drift detection error measures for selected data streams. The scores in bold are statistically related. The bottom row in each measure section shows the average rank for all data stream types.}
    \centering
    \setlength{\tabcolsep}{8pt}
\renewcommand{\arraystretch}{1.2}
\tiny
\begin{tabular}{l|l|rrrrrrr}
\toprule
     & \textsc{ds type} & D3     & KSDD   & OCDD  & PADD   & CDDD  & \textbf{OWADD} & MD3   \\ \midrule
\multirow{7}{*}{\rotatebox[origin=c]{90}{\textsc{D1 measure}}} & N3 | S1.0 & 12.672 & 11.703 & \textbf{9.709} & 11.366 & \textbf{4.393} & \textbf{7.285} & \textbf{11.044} \\
 & N3 | S2.0 & 11.712 & 10.838 & \textbf{6.099} & 10.572 & \textbf{3.116} & \textbf{4.993} & \textbf{6.976} \\
 & N3 | S3.0 & 11.853 & 9.539  & \textbf{6.330}  & 10.146 & \textbf{2.184} & \textbf{4.107} & \textbf{4.301} \\
 & N7 | S1.0 & 6.293  & 6.073  & \textbf{5.110}  & \textbf{5.533} & \textbf{2.648} & \textbf{3.870}  & \textbf{5.621} \\
 & N7 | S2.0 & 6.055  & 5.341  & \textbf{3.908} & 5.036  & \textbf{1.628} & \textbf{2.874} & \textbf{3.695} \\
 & N7 | S3.0 & 5.954  & 4.797  & \textbf{3.608} & 4.532  & \textbf{1.453} & \textbf{2.097} & \textbf{2.403} \\  \cmidrule{2-9}
 & Average rank  & 6.578  & 5.533  & 3.617 & 5.289  & \textbf{1.344} & 2.278 & 3.361 \\\midrule

\multirow{7}{*}{\rotatebox[origin=c]{90}{\textsc{D2 measure}}} &  N3 | S1.0 & 8.099 & 8.444  & 26.704 & \textbf{3.762} & 9.77   & 12.096 & \textbf{5.17}  \\
 & N3 | S2.0 & \textbf{6.513} & 9.019  & 14.11  & \textbf{2.626} & 8.354  & \textbf{5.106} & \textbf{3.886} \\
 & N3 | S3.0 & \textbf{4.981} & 9.421  & 14.435 & \textbf{2.294} & 8.394  & \textbf{3.527} & \textbf{5.471} \\
 & N7 | S1.0 & 8.808 & 9.07   & 25.735 & \textbf{4.026} & 17.601 & 11.755 & \textbf{5.416} \\
& N7 | S2.0 & 8.048 & 9.116  & 12.578 & \textbf{3.41}  & 15.345 & 6.075  & \textbf{4.867} \\
 & N7 | S3.0 & 6.568 & 8.961  & 11.433 & \textbf{2.706} & 14.545 & \textbf{3.692} & \textbf{4.982} \\ \cmidrule{2-9}
& Average rank & 4.006 & 5.117 & 5.825 & \textbf{1.35} & 5.569 & 3.364 & 2.769 \\ \midrule

\multirow{7}{*}{\rotatebox[origin=c]{90}{\textsc{R measure}}} & N3 | S1.0 & \textbf{0.557} & \textbf{0.546} & \textbf{0.721} & 0.743 & 0.958 & \textbf{0.293} & \textbf{0.517} \\
 &N3 | S2.0 & \textbf{0.594} & \textbf{0.515} & \textbf{0.414} & 0.736 & 0.727 & \textbf{0.197} & \textbf{0.284} \\
&N3 | S3.0 & \textbf{0.659} & \textbf{0.497} & \textbf{0.451} & 0.67  & 0.636 & \textbf{0.168} & \textbf{0.208} \\
&N7 | S1.0 & \textbf{0.288} & \textbf{0.288} & 1.925 & \textbf{0.559} & 2.563 & \textbf{0.8}   & \textbf{0.213} \\
&N7 | S2.0 & \textbf{0.278} & \textbf{0.305} & 0.642 & \textbf{0.536} & 1.818 & \textbf{0.367} & \textbf{0.229} \\
 &N7 | S3.0 & \textbf{0.337} & \textbf{0.332} & 0.733 & \textbf{0.49}  & 1.768 & \textbf{0.191} & \textbf{0.295} \\ \cmidrule{2-9}
& Average rank & 4.017 & 3.55 & 4.069 & 5.539 & 5.878 & \textbf{2.439} & \textbf{2.508} \\
\bottomrule
\end{tabular}

    \label{tab:e3_all}
\end{table}

The results show that the proposed \textsc{owadd} is compatible with state-of-the-art unsupervised drift detectors. When evaluating the specific types of streams independently, it is often among the statistically best methods. When statistically comparing the methods across all data streams, it shows the best results in the R measure, statistically related to \textsc{md3}. In the case of the D1 and D2 measures, \textsc{owadd} ranks second and third.

\subsubsection{Novelty recognition}

\begin{figure}[!htb]
    \centering
    \includegraphics[width=0.9\linewidth]{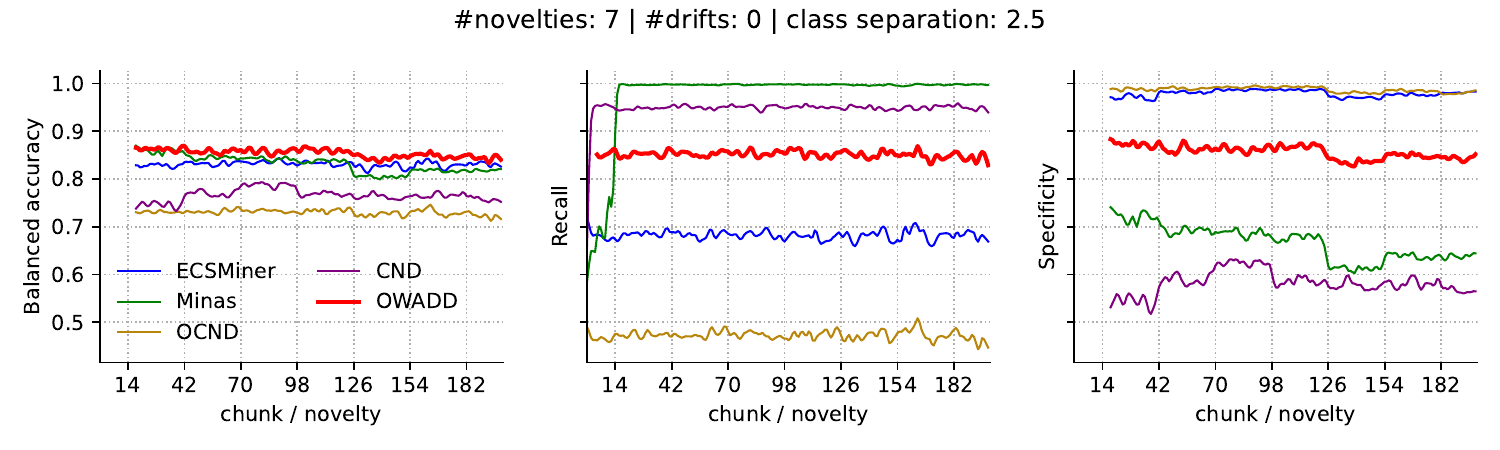}
    \caption{Classification metrics for the task of recognizing novel instances. The results for the proposed method are shown with a bold red line. The ticks on the x-axis indicate the moments of novelty appearance.}
    \label{fig:e4_example}
\end{figure}

The second experiment evaluated the ability of \textsc{owadd} to recognize instances of unknown classes. The study employed three classification metrics to evaluate this task, which was formulated as an imbalanced binary classification problem. The averaged results for data streams with seven novel classes and relatively significant separation of class clusters (2.5) are shown in Figure \ref{fig:e4_example}.

The presented results also highlight the importance of multi-criteria analysis. The proposed method yields the highest results in balanced accuracy, however, when considering recall and specificity independently, it ranks as the third method, being surpassed by \textsc{minas} and \textsc{cnd} in recall and ECSMiner and \textsc{ocdd} in specificity. The subfigure showing the results of specificity (last column) is particularly interesting, as it demonstrates how the appearance of specific novel clusters impacts the overall metrics. For example, the unknown class distribution observed to be the most numerous between chunks 126 and 154 was the most challenging to recognize, as the drop in metrics is observed in almost all studied methods.

\begin{figure}
    \centering
    \includegraphics[width=0.9\linewidth]{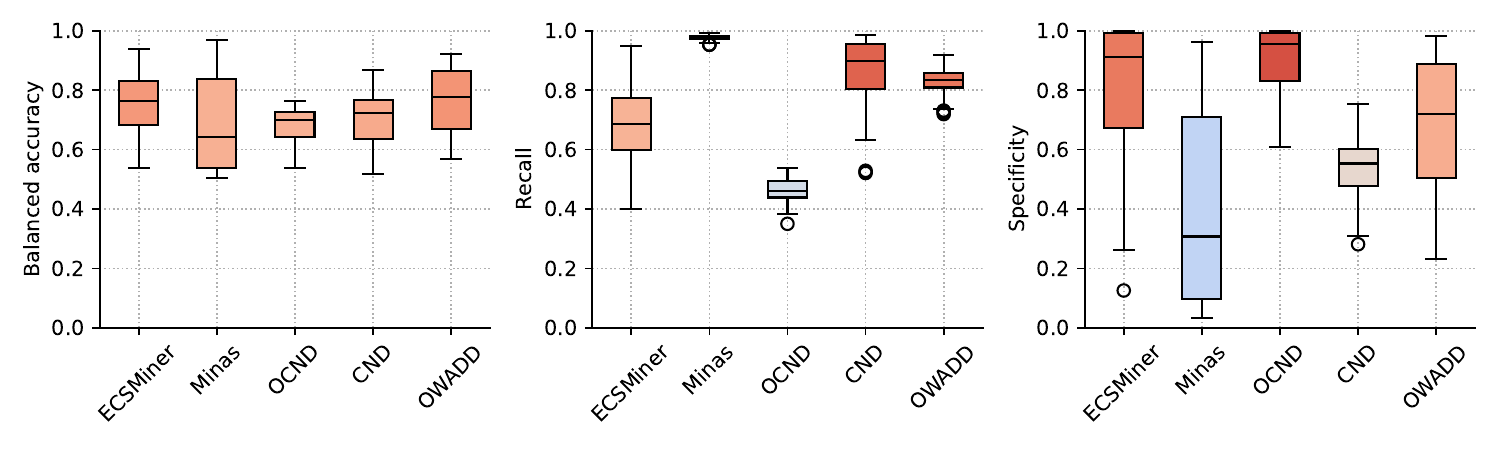}
    \caption{The results for all studied data stream types in the task of novelty recognition. The color of a box indicates the average result of a method, with a higher score closer to red.}
    \label{fig:e4_all}
\end{figure}

Figure \ref{fig:e4_all} graphically presents the distributions of all obtained results for the reference and proposed \textsc{owadd} approaches. The subfigures present three studied metrics. The results again show that high quality in a specific metric often translates to poor recognition ability across another. This can be seen in the results of \textsc{ocnd}, which has excellent specificity but the lowest recall, indicating difficulty in recognizing known class instances. Such a combination of recall and specificity translated to a relatively low balanced accuracy. In contrast, the results of the proposed \textsc{owadd} show a good balance between recall and specificity, achieving high balanced accuracy.

\begin{table}[!b]
\caption{Classification metric for novelty recognition. The scores in bold are statistically related. The bottom row shows the average rank for all data stream types.}
    \centering
    \setlength{\tabcolsep}{13pt}
\renewcommand{\arraystretch}{1.2}
\tiny
\begin{tabular}{l|l|rrrrr}
\toprule
      & \textsc{ds type}  & ECSMiner & Minas & OCND  & CND   & \textbf{OWADD} \\ \midrule
\multirow{7}{*}{\rotatebox[origin=c]{90}{\textsc{balanced accuracy}}} & N5 | S1.0 & \textbf{0.601} & 0.513 & \textbf{0.587} & \textbf{0.583} & \textbf{0.614} \\

& N5 | S2.0 & \textbf{0.795} & 0.649 & \textbf{0.711} & \textbf{0.738} & \textbf{0.775} \\
& N5 | S3.0 & \textbf{0.841} & \textbf{0.885} & \textbf{0.736} & \textbf{0.774} & \textbf{0.882} \\
& N9 | S1.0 & \textbf{0.623} & 0.513 & \textbf{0.602} & \textbf{0.584} & \textbf{0.622} \\
& N9 | S2.0 & \textbf{0.778} & 0.641 & \textbf{0.707} & \textbf{0.733} & \textbf{0.763} \\
& N9 | S3.0 & \textbf{0.826} & \textbf{0.926} & \textbf{0.739} & \textbf{0.748} & \textbf{0.899} \\ \cmidrule{2-7}
& Average rank & \textbf{3.767} & 2.333 & 2.027 & 2.687 & \textbf{4.187} \\ \midrule

\multirow{7}{*}{\rotatebox[origin=c]{90}{\textsc{recall}}} & N5 | S1.0 & 0.731 & \textbf{0.97} & 0.448 & 0.695 & \textbf{0.808} \\
& N5 | S2.0 & 0.664 & \textbf{0.968} & 0.476 & \textbf{0.902} & \textbf{0.823} \\
& N5 | S3.0 & 0.686 & \textbf{0.97} & 0.48  & \textbf{0.969} & \textbf{0.843} \\
& N9 | S1.0 & 0.735 & \textbf{0.979} & 0.472 & 0.698 & \textbf{0.831} \\
& N9 | S2.0 & 0.625 & \textbf{0.982} & 0.47  & \textbf{0.909} & \textbf{0.838} \\
& N9 | S3.0 & 0.657 & \textbf{0.98} & 0.481 & \textbf{0.958} & \textbf{0.852}\\ \cmidrule{2-7}
& Average rank & 2.227 & \textbf{4.913} & 1.047 & 3.593 & 3.22 \\ \midrule

\multirow{7}{*}{\rotatebox[origin=c]{90}{\textsc{specificity}}} & N5 | S1.0 & 0.473 & 0.058 & \textbf{0.727} & 0.471 & 0.421 \\
& N5 | S2.0 & \textbf{0.927} & 0.332 & \textbf{0.945} & 0.575 & 0.726 \\
& N5 | S3.0 & \textbf{0.996} & 0.802 & \textbf{0.991} & 0.579 & \textbf{0.921} \\
& N9 | S1.0 & 0.511 & 0.048 & \textbf{0.732} & 0.471 & 0.413 \\
& N9 | S2.0 & \textbf{0.932} & 0.301 & \textbf{0.943} & 0.558 & 0.689 \\
& N9 | S3.0 & \textbf{0.996} & 0.871 & \textbf{0.997} & 0.538 & \textbf{0.946} \\ \cmidrule{2-7}
& Average rank & \textbf{4.15} & 1.373 & \textbf{4.637} & 1.933 & 2.907 \\
\bottomrule
\end{tabular}

    \label{tab:e4}
\end{table}

\begin{figure}[!htb]
    \centering
    \includegraphics[width=0.32\linewidth, trim={0.8cm 0.2cm 0.8cm 0.2cm}, clip]{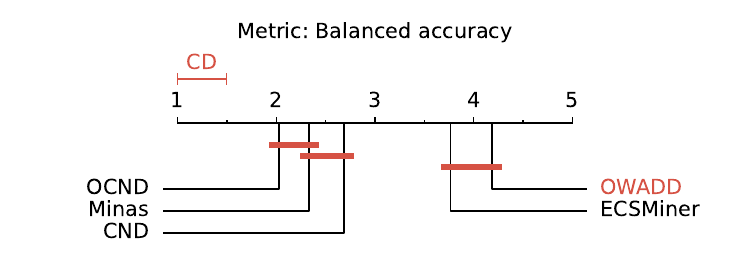}
    \includegraphics[width=0.32\linewidth, trim={0.8cm 0.2cm 0.8cm 0.2cm}, clip]{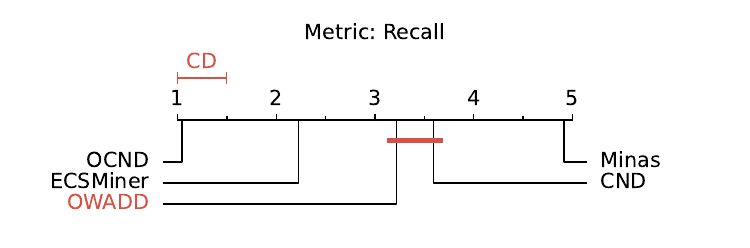}
    \includegraphics[width=0.32\linewidth, trim={0.8cm 0.2cm 0.8cm 0.2cm}, clip]{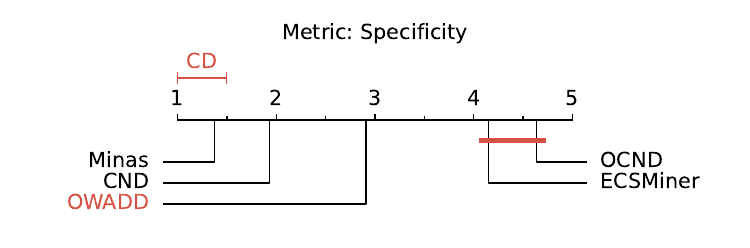}
    \caption{Critical difference diagrams for the results of the novelty recognition experiment.}
    \label{fig:e4_cd}
\end{figure}

The detailed results of this experiment for the selected data streams are presented in Table \ref{tab:e4}. Similarly to Table \ref{tab:e3_all}, the cells indicate the average metric value for a specific type of data stream (where $N$ denotes the number of novelties and $S$ the class separation), and the last row in each section shows the average rank across all studied data streams. The results of the statistical analysis using the Friedman test and post-hoc Nemenyi test are also presented in Figure \ref{fig:e4_cd}.

The obtained results show that the proposed \textsc{owadd} method, statistically related to ECSMiner, achieves the highest scores in balanced accuracy metrics. In two other considered metrics, the method ranks third, being outranked by \textsc{minas} and statistically related to \textsc{cnd} in recall, and outranked by \textsc{ocnd} and ECSMiner in specificity. We note that none of the reference methods consistently outperforms the proposition across all evaluated metrics, which highlights that \textsc{owadd} effectively balances its ability to recognize both known classes and novelties in this dynamically imbalanced setting.

\section{Conclusions}

This work proposes the \textsc{owadd} approach for non-stationary data stream processing with concept drifts and novel class appearances. The method uses the reconstruction error of two mirrored fully connected autoencoders to (1) detect concept drifts using statistical analysis of autoencoding errors, and (2) recognize unknowns by thresholding the scores after estimating error density. During the initial training, the weights of both autoencoders are optimized together, but they operate independently for the tasks of drift detection and novelty recognition, allowing for appropriate adaptation to changing problem distributions. The use of a one-dimensional reconstruction error \textit{proxy}, as an alternative to the analysis of the entire feature space, limits the memory complexity of the method.The research compares the proposed method to state-of-the-art techniques on two tasks: unsupervised drift detection, where the appearance of novel classes is also considered a form of concept drift, and novelty recognition with an increasing number of unknown class samples.

Future work will focus on adapting the method to unstructured data, where, instead of a fully connected architecture, the \textsc{owadd} could utilize a convolutional autoencoder, allowing for the recognition of concept changes in computer vision tasks. The presented experimental evaluation focused on synthetic data with sudden concept drifts. Future research will extend the analysis to gradual and incremental changes and consider the recurrence of concepts.

\begin{credits}
\subsubsection{\ackname} 
This work was supported by the statutory funds of the Department of Systems and Computer Networks, Faculty of Information and Communication Technology, Wrocław University of Science and Technology.

\subsubsection{\discintname}
The authors have no competing interests to declare that are
relevant to the content of this article.
\end{credits}

\bibliographystyle{splncs04}
\bibliography{bibliography}

\end{document}